\pdfoutput=1

\documentclass[11pt]{article}

\usepackage[final]{acl}
\usepackage{amsmath}
\usepackage{times}
\usepackage[most]{tcolorbox}
\usepackage{xspace}
\usepackage{array}
\usepackage{amsfonts}
\usepackage{multirow} 
\usepackage[ruled,vlined]{algorithm2e}
\usepackage{caption}
\usepackage{booktabs}
\usepackage{longtable}
\usepackage{enumitem}
\usepackage{multicol}
\usepackage[T1]{fontenc}

\usepackage[utf8]{inputenc}

\usepackage{microtype}

\usepackage{inconsolata}

\usepackage{graphicx}
\newcommand{\our}{\textsc{zoes}\xspace}

\newcommand{\lb}{$\langle$}
\newcommand{\rb}{$\rangle$}

\title{\textbf{Zero-Shot Open-Schema Entity Structure Discovery}}

\author{
\textbf{Xueqiang Xu\textsuperscript{1}},
\textbf{Jinfeng Xiao\textsuperscript{2}}\thanks{Prior to the co-author's role at Amazon},
\textbf{James Barry\textsuperscript{4}},
\textbf{Mohab Elkaref\textsuperscript{4}},\\
\textbf{Jiaru Zou\textsuperscript{1}},
\textbf{Pengcheng Jiang\textsuperscript{1}},
\textbf{Yunyi Zhang\textsuperscript{1}},
\textbf{Max Giammona\textsuperscript{3}},\\
\textbf{Geeth de Mel\textsuperscript{4}},
\textbf{Jiawei Han\textsuperscript{1}} \\
\\
\textsuperscript{1}University of Illinois at Urbana-Champaign \\
\textsuperscript{2}Amazon \quad
\textsuperscript{3}IBM Research \quad
\textsuperscript{4}IBM Research Europe\\
\texttt{\{xx19, hanj\}@illinois.edu} 
}

\begin{document}
\maketitle

\begin{abstract}
Entity structure extraction, which aims to extract entities and their associated attribute–value structures from text, is an essential task for text understanding and knowledge graph construction. Existing methods based on large language models (LLMs) typically rely heavily on predefined entity attribute schemas or annotated datasets, often leading to incomplete extraction results. To address these challenges, we introduce \textit{Zero-Shot Open-schema Entity Structure Discovery} (\our), a novel approach to entity structure extraction that does not require any schema or annotated samples. \our operates via a principled mechanism of enrichment, refinement, and unification, based on the insight that an entity and its associated structure are mutually reinforcing. Experiments demonstrate that \our consistently enhances LLMs' ability to extract more complete entity structures across three different domains, showcasing both the effectiveness and generalizability. These findings suggest that such an enrichment, refinement, and unification mechanism may serve as a principled approach to improving the quality of LLM-based entity structure discovery in various scenarios.
\end{abstract}
\section{Introduction}\label{intro}
\begin{figure}[htbp]
  \centering
  \includegraphics[width=\columnwidth]{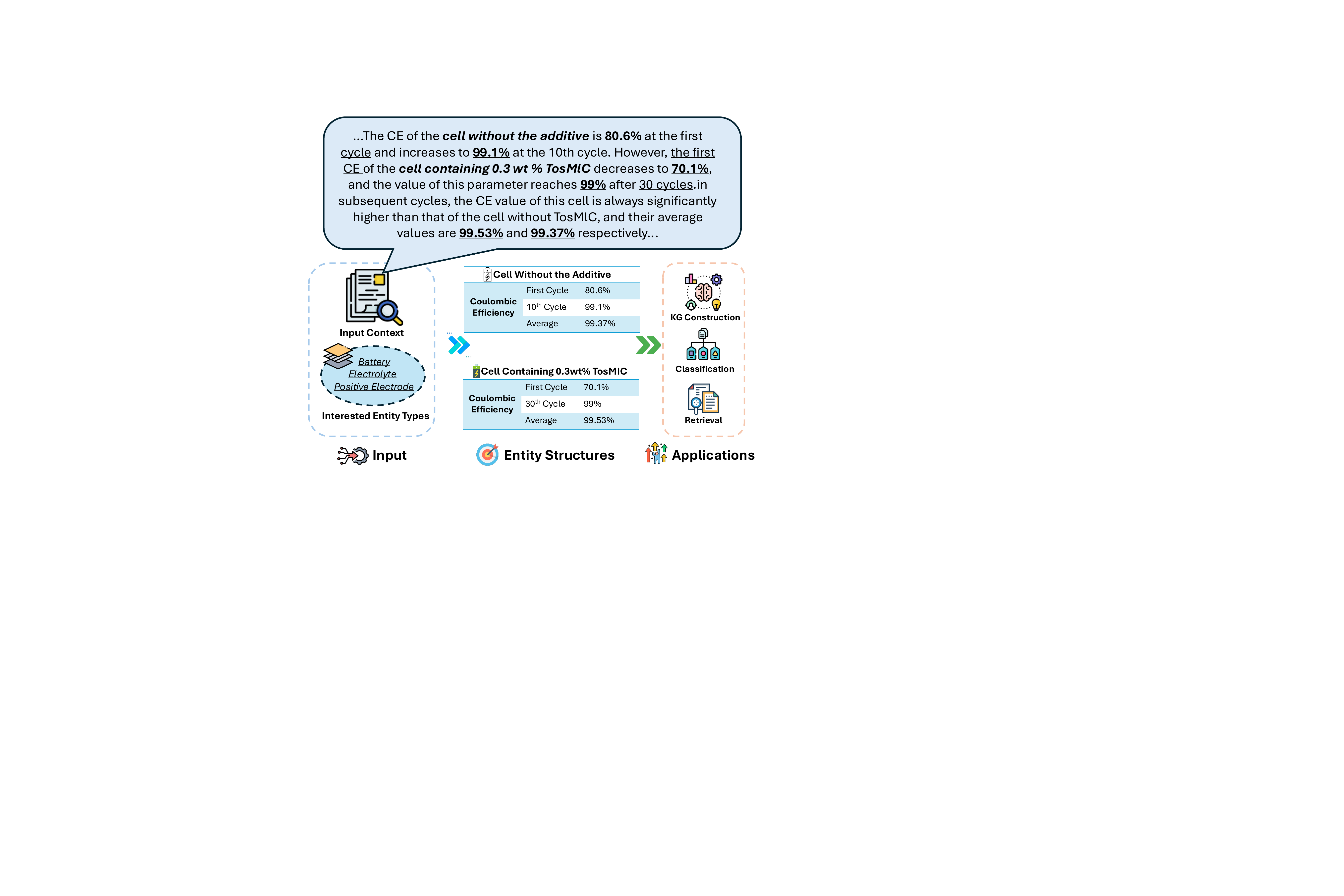}
  \caption{An example of the entity structure discovery task with applications.  
           The figure depicts CEs of two discovered cells under different conditions organized as in the source passage from~\citep{task_demonstration}.}
  \label{fig:task-overview}
\end{figure}
Automatic mining of structured entity information is critical for knowledge discovery and management~\citep{kg_survey, arsenyan-etal-2024-largeKG}. Prior works on entity information extraction—including relation extraction~\citep{ding-etal-2024-priore, zhou2024relation_extraction, zhang-etal-2025-ce-RE}, entity typing~\citep{Onoe_Durrett_2020_typing, tong-etal-2025-evoprompt_typing}, and named entity recognition~\citep{ner2020survey, keraghel2024ner_survey}—have primarily focused on extracting isolated aspects of entity knowledge. However, modeling only a single aspect of entity information may be insufficient for real-world applications~\citep{jiao-etal-2023-instruct, Dagdelen2024}. For example, in the battery science domain, a battery’s performance is determined by complex conditions~\cite{zhou2023corpus}. As shown in Figure~\ref{fig:task-overview}, even for the same battery, its ``Coulombic Efficiency'' (CE) value varies across different cycles. A single triplet (e.g., \lb Cell Without the Additive, CE at First Cycle, 80.6\% \rb) conveys limited information about the battery’s performance. In contrast, unifying performance across different conditions into a structured representation provides a clearer and more comprehensive view. Therefore, there is a need for a unified representation of entity information that integrates multiple aspects rather than focusing on a single one~\citep{lu-etal-2023-pivoine}.

Recently, closed-schema entity structure extraction has been proposed to unify various aspects of entity information under predefined type schemas, where each entity type is associated with a fixed set of attributes~\citep{zhong-etal-2023-reactie, wu-etal-2024-structured_entity}. The goal is to extract structured entities represented as the entity along with a set of $\langle \text{attribute}, \text{value} \rangle$ pairs. By combining with the entity name to form $\langle \text{entity}, \text{attribute}, \text{value}\rangle$ triplet, it can capture a specific property of the entity, as illustrated in Figure~\ref{fig:task-overview}. However, like other closed-schema information extraction tasks~\citep{li2021document, 10.1145/3488560.3498377, zhou2023corpus}, entity type schemas confine the extraction on a limited set of attributes, which fail to capture diverse and unseen attributes in fast-evolving real-world scenarios~\citep{openIE-2024-survey}.

To enable entity structure extraction to capture more diverse and dynamic information, we extend traditional closed-schema entity structure extraction to an open information extraction setting~\citep{openie_2016}, which we term \textbf{Open-Schema Entity Structure Discovery} (OpenESD). In OpenESD, we want to identify entities within user interests \emph{and} their $\langle \text{attribute}, \text{value} \rangle$ structures \emph{without} any predefined attribute sets as a schema. OpenESD can benefit many downstream tasks such as information retrieval~\citep{kang-etal-2024-taxonomy} and question answering~\citep{edge2025localglobalgraphrag, gutiérrez2025ragmemorynonparametriccontinual, jiang2025ras}. With an open-schema setting, OpenESD goes beyond straightforward extraction: it demands discovering~\citep{jiao-etal-2023-instruct, pei-etal-2023-abstractive}, organizing~\citep{wu-etal-2024-structured_entity}, and inferring~\citep{ding-etal-2024-priore} the most appropriate attributes and values for each entity. 

Large language models (LLMs) with extensive parametric knowledge have demonstrated promising performance in open information extraction~\citep{jiao-etal-2022-open, lu-etal-2023-pivoine}, offering a promising solution for OpenESD. However, fully harnessing this capability remains challenging.
(i) \textbf{Extraction Coverage}: An LLM tends to capture coarse-grained facts that are more frequent in its parametric knowledge while missing rare, fine-grained information from the context.
(ii) \textbf{Extraction Granularity}: When the context contains rich details, LLMs may fail to identify the appropriate level of granularity for representing the extracted information, resulting in incomplete or ambiguous structures. For example, as illustrated in Figure~\ref{fig:task-overview}, if the extracted ``CE'' attributes fail to capture contextual conditions, multiple ``CE'' values may be incorrectly mapped to the same attribute, leading to inaccurate results.

To enhance LLMs' capability on OpenESD, we introduce \our, a \textbf{\underline{z}ero-shot \underline{o}pen-schema \underline{e}ntity \underline{s}tructure discovery} framework. By employing a principled mechanism of enrichment, refinement, and unification, \our effectively extracts structured entity information without supervision. Specifically, \our starts with the LLM's zero-shot $\langle \text{entity}, \text{attribute}, \text{value} \rangle$ triplets results, then gradually discovers new triplets to enrich it. Next, \our leverages mutual dependencies among triplet elements to identify and refine inferior triplets. 
Finally, the refined triplets are aggregated into entity structures as coherent representations of the entities based on user interest.

We evaluate \our using different backbone models on one long-tail domain: \texttt{Battery Science} and two general domains: \texttt{Economics} and \texttt{Politics}. The results demonstrate that \our can consistently outperform baselines with different backbone models in all domains. \our achieves an absolute improvement of $+10.64\%$ in the F1 score. These results demonstrate the effectiveness and generalizability of our method for OpenESD.

Our contributions are summarized as follows: 
\begin{enumerate}[leftmargin=*,nosep]
\item We introduce open-schema entity structure discovery, a task to automatically identify entities within user interest along with their contextual $\langle \text{attribute}, \text{value} \rangle$ structures without any predefined schema, which can benefit several knowledge intensive tasks.

\item We propose \our, a zero-shot open-schema entity structure discovery method. By the enrich-refine-unify strategy, \our substantially improves LLMs' performance on OpenESD.

\item We evaluate \our and baselines on three very different domains to further study LLMs' capabilities on OpenESD.
\end{enumerate}


\section{Related Work}
\subsection{Open Information Extraction}
Open Information Extraction (OpenIE) aims to extract structured information from unstructured text without relying on predefined schemas~\citep{Zheng_2018, openIE2022, openIE-2024-survey}. Early OpenIE relied on rule-based methods~\citep{clauseIE, openie_2016}, sequence labeling~\citep{ro-etal-2020-multi,Vasilkovsky2022detie,yu-etal-2021-maximal}, or sequence-to-sequence models~\citep{kolluru-etal-2022-alignment} to extract relational triplets from individual sentences. However, sentence-level relation extractions cannot capture cross-sentence relational information~\citep{dunn2022structuredinformationextractioncomplex, wu-etal-2024-structured_entity}, which leads to low information extraction coverage~\citep{li2021document, Dagdelen2024}. 

Recent advances in OpenIE focus on leveraging LLMs to perform more expressive and instruction-following extractions~\citep{jiao-etal-2023-instruct,qi-etal-2024-adelie}. \citet{pei-etal-2023-abstractive} demonstrates that this paradigm can identify more triplets whose predicates are not explicitly mentioned. These LLMs support more flexible and user-guided information extraction, moving beyond fixed triplet formats toward on-demand schemas~\citep{pei-etal-2023-abstractive,qi-etal-2024-adelie}. While these approaches significantly improve the coverage and adaptability of OpenIE, they typically require substantial annotated training data or task-specific instruction tuning~\citep{lu-etal-2023-pivoine}, which constrains their applicability in low-resource or specialized domains~\citep{Wei2023ZeroShotIE}. Compared with previous works, \our focuses on a generalizable approach to guide LLMs to unify document-level entity information into structured representations by leveraging internal structural consistency, rather than relying on extensive training or annotations.
\subsection{Zero-shot Relation Extraction}
Zero-shot relation extraction (ZSRE) aims to identify semantic relations between entities without relying on labeled training instances~\citep{levy-etal-2017-zero}. Prior work has predominantly approached this task by leveraging semantic representations to generalize to unseen relations~\citep{chen-li-2021-zs, Tran2022ImprovingDL, zhao-etal-2023-matching}. For example, \citet{chen-li-2021-zs} proposed ZS-BERT, a supervised model that learns relation embeddings from attribute descriptions. Similarly, \citet{zhao-etal-2023-matching} introduced a fine-grained matching framework that integrates both entity and context embeddings to enhance zero-shot prediction. However, such embedding-based methods are sensitive to the exact wording of relation labels, limiting their robustness and generalizability in real-world settings.

More recently, LLMs have enabled a new paradigm in zero-shot relation extraction~\citep{li2023revisiting, xue2024autore, zhou2024relation_extraction, li-etal-2025-frame}. One line of work explores using LLMs to generate relational statements directly from entity mentions, rather than extracting from predefined relation schemas or sentence-level contexts~\citep{jiang-etal-2024-genres, ding-etal-2024-priore}. For instance, \citet{ding-etal-2024-priore} leverage LLMs' understanding of entity types to generate topic-specific relations by aggregating corpus-level evidence. While these methods demonstrate strong generalization capabilities, they often produce high-level or generic relations. Our work explores utilizing LLMs to extract contextualized entity structures directly from input context without external knowledge.

\section{Method}
\begin{figure*}[t]               
  \centering
  \includegraphics[width=\textwidth]{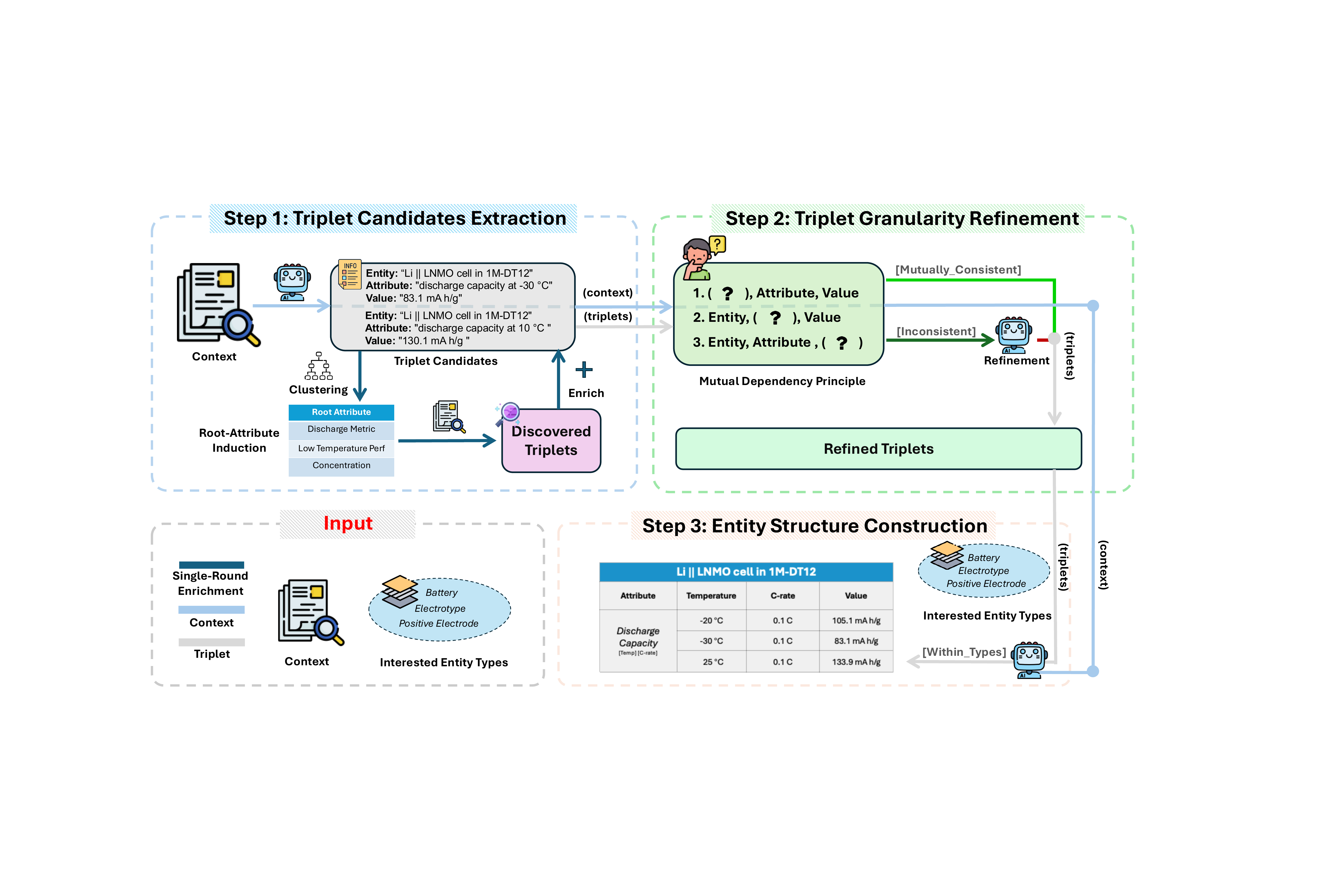}
  \caption{\textbf{Methodology Overview of \our.} \our operates in three stages: (1) \textbf{Triplet Candidates Extraction} expands the initial zero-shot EAV triplet set by leveraging generalized root attributes induced from initial extractions as guidance to uncover additional triplets; (2) \textbf{Triplet Granularity Refinement} applies the triplet mutual dependency principle to detect and revise under-specified or inconsistent triplets; and (3) \textbf{Entity Structure Construction} assembles refined triplets into entity structures, which are filtered based on user-specified target entity types.}
  \label{fig:method-overview}
\end{figure*}

In this section, we start with the task formulation of open-schema entity structure discovery, and then delve into \our, a three-stage approach for performing the task of OpenESD in detail. An illustrated overview of \our is in Figure~\ref{fig:method-overview}.

\subsection{Task Formulation}
Open-schema entity structure discovery aims to automatically identify entities and their corresponding structures, from an input document and a given set of entity types of interest, without relying on any pre-defined schemas (e.g., pre-defined attribute names). 
The structure of each entity is represented as a set of $\langle$attribute, value$\rangle$ pairs, where entities and their associated structures are derived from the document. 
As an example, Figure~\ref{fig:task-overview} contains a battery science domain document discussing multiple properties regarding the entities ``Cell Without the Additive'' and ``Cell Containing 0.3wt\% TosMIC''. The discovered entity structures should organize those properties as a set of attribute-value pairs, like attribute: ``CE at First Cycle'' with value: ``80.6\%'' for ``Cell Without the Additive''.

Formally, given a document $d$ and a set of entity types of interest $\mathcal{T}$, the goal is to identify a set of entities $\mathcal{E}$ within $\mathcal{T}$ such that $\mathcal{E} = \{e_1, \ldots, e_{m}\}$ and extract the structure of each entity. 
For an entity $e_i \in \mathcal{E}$, let
$A_i = \{a_{i,1}, \ldots, a_{i,n_i}\}$ be the set of attributes
and $V_i = \{v_{i,1}, \ldots, v_{i,n_i}\}$ be the corresponding set of values. 
We then define the structure $S_i$ as 
\[
  \mathcal{S}_i = \bigl\{\, (a_{i,j},v_{i,j}) \,\mid\, j \in \{ 1, \dots, n_i \} \bigr\}.
\]

\subsection{Triplet Candidates Extraction}
Zero-shot triplet extraction using LLMs often suffers from limited knowledge coverage, as LLMs tend to prioritize extracting explicitly mentioned and high-frequency attribute-value pairs. \citet{edge2025localglobalgraphrag} attempt to improve coverage by prompting LLMs for multiple extraction rounds. However, without targeted guidance, such multi-round generation frequently yields redundant or noisy triplets, while still failing to recover low-salience but semantically meaningful triplets.

To address this challenge, \our first induces root attributes from an LLM's initial extracted triplets $T_{initial}$. These root attributes serve as semantic guidance that clarify what kinds of values are valid or expected from the context, which assists the LLM to revisit the document context to discover missing triplets.

\paragraph{Root Attribute Induction.}
The initial zero-shot extraction yields a set of $\langle \text{entity}, \text{attribute}, \text{value} \rangle$ triplets $T_{initial}$, where some attributes are specific (e.g., \textit{``CE at first cycle''}, \textit{``initial CE''}). Such fine-grained attributes often correspond to only one triplet. In contrast, a general attribute such as \textit{``Coulombic Efficiency''} can map to a set of potential values. We can utilize more general attributes to identify those previously missing values, thus identifying missing triplets.

Motivated by this observation, we induce \textit{root attributes} that abstract over semantically similar attributes to guide the subsequent triplet enrichment stage. Specifically, we first map all extracted attributes into a latent space using a dense encoder \citep{wang2022text}. To group attributes expressing the same underlying concept, we apply agglomerative clustering \citep{ward1963hierarchical} using a cosine distance metric:
\begin{equation}
    d(a_m, a_n) = 1 - \cos(\text{Enc}(a_m), \text{Enc}(a_n))
\end{equation}
where $a_m$ and $a_n$ denote individual attributes. Finally, for each resulting cluster, we prompt an LLM to synthesize a coarse-grained root attribute from its members (e.g., \textit{``Coulombic Efficiency''} derived from \textit{``CE at first cycle''} and \textit{``CE at the 10th cycle''}).

\paragraph{Value-Anchored Enrichment.}\label{triplet_enrichment}
Once root attributes are identified, we use them to guide the discovery of additional value mentions. For each root attribute, we prompt the LLM to revisit the document and list all corresponding values. This step often recovers contextually grounded values (e.g., \textit{``higher''} a value comparing the CE among two cells) that align with the root attribute but were missed initially.

Although some entities may lack explicitly stated attribute–value structures in the context, each semantically meaningful value (e.g., "80.6\%") should correspond to at least one valid triplet. Based on this intuition, each newly discovered value is treated as an anchor to elicit a missing triplet. We then prompt the LLM to infer the corresponding entity and attribute, constrained by the associated root attribute. This targeted prompting enables the recovery of under-expressed or indirectly stated facts, significantly improving extraction coverage.

By using root attributes as interpretable guides and values as anchors, this enrichment process helps the LLM uncover a more complete and semantically coherent $\langle\text{entity}, \text{attribute}, \text{value}\rangle$ triplet set $T_{enrich}$.

\subsection{Triplets Granularity Refinement}\label{method: refinement}
Directly prompting LLMs to produce triplets in a zero-shot setting often yields suboptimal results to capture complex conditions, since LLMs lack an explicit understanding of the granularity required to represent entity structures unambiguously. To address this, we propose a refinement mechanism grounded in the \textbf{Mutual Dependency Principle}:
\begin{quote}
    {For a triplet \( t = \langle e, a, v \rangle \), we assume that appropriate granularity is achieved when any one component can be reliably inferred from the other two within the context \( d \).}
\end{quote}
Based on this principle, given a triplet \( t = \langle e, a, v \rangle \) from context \( d \), we generate three questions, each aiming to recover one component based on the other two and the context.
Specifically, for each triplet \( t_i = \langle e_i, a_i, v_i \rangle \in T_{\text{enrich}} \), we construct:
\begin{align*}
QA(t_i, d) = \big\{\, \langle q_c, \text{ans}_c \rangle \;\big|\, 
& \text{ans}_c \in \{e_i, a_i, v_i\}, \\
&\hspace{-1.7em} q_c \in \text{LLM}(t_i, d)
\,\big\}
\end{align*}
For example, for a triplet \lb Cell without the Additive, CE, higher \rb, we can construct questions:
\begin{itemize}[leftmargin=*, nosep]
    \item \textit{Which cell shows a higher CE?}
    \item \textit{What is higher for the cell without the additive?}
    \item \textit{What is the CE of the cell without the additive?}
\end{itemize}
The LLM is then prompted to answer these questions based on context $d$. We compare the predicted answer \( ans_p \) with the masked ground-truth component \( ans_c \). A triplet is considered \textbf{mutually consistent} if all three components can be accurately recovered. Otherwise, it is flagged for refinement. For instance, if the original triplet is \lb Cell without the Additive, CE, higher \rb, by giving only the entity and attribute, multiple values can be inferred from the context, which are not necessarily ``higher''. This indicates that the attribute lacks specificity and needs refinement. To perform refinement, we treat the value \( v_i \) as an anchor and prompt the LLM to revise the entity and attribute conditioned on \( v_i \) and context $d$.

This dependency-driven refinement helps identify and correct coarse or under-specified triplets, ensuring that only mutually-consistent triplets are retained. We denote the final set of refined triplets as $T_{refine}$, which serves as the input to the subsequent structure construction phase. 


\subsection{Entity Structure Construction}
The final step of \our is to merge refined triplets into coherent entity structures, as illustrated in Figure~\ref{fig:task-overview}. Since the refinement step (Section~\ref{method: refinement}) utilizes the mutual dependency principle, the resulting triplets possess better granularity to accurately convey meaningful information unambiguously. To construct entity structures, we directly prompt the LLM with both the document context \( d \) and the refined triplet set \( T_{\text{refine}} \) to merge triplets discussing the same entity to form entity structures \( \mathcal{E}_{\text{initial}} \).

\paragraph{Structure-Aware Filtering} In real-world applications, users often have specific types of entities of interest, denoted as a target type set \( \mathcal{T} \). For each structured entity \( e_i \in \mathcal{E}_{\text{initial}} \), we use the LLM to determine whether it belongs to the desired types, based on its attributes, values, and the document context:
\[
\text{LLM}(e_i, \mathcal{T} \mid d) \rightarrow \{\text{True}, \text{False}\}
\]
This structure-aware filtering enables \our to utilize entity structures to augment entity names' semantics. In many domains, entity names alone are insufficient to determine their relevance or type. For instance, in battery science, entities such as ``fluoroethylene carbonate'' may not clearly indicate its entity types even with context. However, if we know it has an attribute as a function in battery electrolyte, the LLM can directly know its type is ``electrolyte additive''. Finally, by construction and filtration, \our can produce contextually grounded entity structures $\mathcal{E}$ in a zero-shot setting.

\section{Experiments}
\begin{table*}[htbp]
\centering
\small
\setlength{\tabcolsep}{6pt} 
\renewcommand{\arraystretch}{1.2} 
\resizebox{\textwidth}{!}{%
\begin{tabular}{>{\centering\arraybackslash}m{2cm} >{\centering\arraybackslash}m{3.1cm} |ccc |ccc |ccc}
\toprule
\multirow{2}{*}{\textbf{Model}} & \multirow{2}{*}{\textbf{Method}} & \multicolumn{3}{c}{\textbf{Battery Science}} & \multicolumn{3}{|c}{\textbf{Economics}} & \multicolumn{3}{|c}{\textbf{Politics}} \\
\cmidrule(lr){3-5} \cmidrule(lr){6-8} \cmidrule(lr){9-11}
& & \textbf{Precision} & \textbf{Recall} & \textbf{F1} & \textbf{Precision} & \textbf{Recall} & \textbf{F1} & \textbf{Precision} & \textbf{Recall} & \textbf{F1} \\
\midrule
\textbf{Llama-3.2-3B} 
& Text2Triple (SFT) & 0.2634 & 0.1718 & 0.2083 & 0.7312 & 0.6538 & 0.6903 & 0.8416 & 0.7852 & 0.8124 \\
\midrule
\multirow{3}{*}{\textbf{GPT-4o}} 
& CoT & 0.6087 & 0.4275 & 0.5022 & 0.8880 & 0.6619 & 0.7585 & 0.8214 & 0.1593 & 0.2669 \\
& Few-Shot & \textbf{0.7911} & 0.4771 & 0.5952 & \textbf{0.9046} & 0.7149 & 0.7986 & \textbf{0.9295} & 0.6397 & 0.7579 \\
& Zoes (Ours) & 0.7758 & \textbf{0.6844} & \textbf{0.7287} & 0.8994 & \textbf{0.9104} & \textbf{0.9049} & 0.8534 & \textbf{0.9007} & \textbf{0.8764} \\
\midrule
\multirow{3}{*}{\textbf{GPT-4o-mini}} 
& CoT & 0.5562 & 0.3779 & 0.4500 & 0.8493 & 0.5967 & 0.7010 & 0.5952 & 0.1155 & 0.1934\\
& Few-Shot & 0.5102 & 0.3816 & 0.4367 & \textbf{0.8657} & 0.7352 & 0.7952 & \textbf{0.8933} & 0.6767 & 0.7700 \\
& Zoes (Ours) & \textbf{0.5708} & \textbf{0.6441} & \textbf{0.6104} & 0.8532 & \textbf{0.8289} & \textbf{0.8409} & 0.8374 & \textbf{0.7852} & \textbf{0.8105} \\
\midrule
\multirow{3}{*}{\textbf{Granite-8B}} 
& CoT & 0.6149 & 0.3473 & 0.4439 & 	0.7051 & 0.4236 & 0.5293 & 0.7241 & 0.0970 & 0.1711 \\
& Few-Shot & \textbf{0.6579} & 0.3817 & 0.4831 & 0.7398 & 0.5153 & 0.6074 & 0.7431 & 0.4341 & 0.5481 \\
& Zoes (Ours) & 0.5708 & \textbf{0.5229} & \textbf{0.5458} & \textbf{0.8017} & \textbf{0.7821} & \textbf{0.7918} & \textbf{0.7790} & \textbf{0.8383} & \textbf{0.8076} \\
\bottomrule
\end{tabular}%
}
\caption{Evaluation with user interested entity types across different backbone models and methods on Battery Science, Economics, and Politics. Bold numbers highlight the best results per backbone model in Battery Science.}
\label{tab: results}
\vspace{-0.5em}
\end{table*}

We begin with the experimental setup, including dataset construction, evaluation metrics, and implementation details. We then present our main results, followed by ablation studies evaluating the effectiveness of each component in \our.
\subsection{Dataset Construction}
\begin{table}[t]
\centering
\small
\setlength{\tabcolsep}{6pt} 
\renewcommand{\arraystretch}{1.3}
\begin{tabular}{l|ccc}
\toprule
\textbf{Domain} & \textbf{\#Documents} & \textbf{\#Sentences} & \textbf{\#(E, A, V)s} \\
\midrule
\textbf{BatSci}   & 20  & 197 & 428  \\
\textbf{Economics}  & 50  & 195 & 491  \\
\textbf{Politics} & 50  & 208 & 433  \\
\midrule
\textbf{Overall}  & 120 & 675 & 1,289 \\
\bottomrule
\end{tabular}
\caption{Dataset statistics across ``Battery Science'', ``Economics'', and ``Politics'' domains. ``BatSci'' stands for ``Battery Science,'' and ``\((E, A, V)\)s'' denotes \(\langle \text{entity}, \text{attribute}, \text{value} \rangle\) triplets.}
\label{tab:dataset-statistics}
\vspace{-0.5em}
\end{table}

The aim of OpenESD is to discover entities with their attribute value structures from the context, where attributes are often implicitly hidden~\citep{pei-etal-2023-abstractive}. We construct an entity structure extraction dataset spanning one long-tail domain, \texttt{Battery Science}, and two general domains, \texttt{Economics} and \texttt{Politics}. The dataset specifically focuses on evaluating two challenges of OpenESD: \textit{extraction coverage} and \textit{extraction granularity} introduced in Section~\ref{intro}. For each domain, the dataset contains a set of documents and a set of interested entity types. The statistics of the dataset can be found in Table~\ref{tab:dataset-statistics}.

\paragraph{Long-Tail Domain.} For the \texttt{Battery Science} domain, we curate paragraphs from top-tier peer-reviewed research articles that discuss the performance and applications of battery components. These paragraphs are characterized by diverse experimental conditions and frequent comparisons across similar components. Missing contextual conditions in such cases can result in misleading or contradictory information. Furthermore, the text contains domain-specific terminology and fine-grained technical descriptions, posing significant challenges for LLMs to accurately understand and extract entity structures. This domain exemplifies the long-tail scenario: high knowledge granularity, low representation in pretraining corpora, and substantial variance in how attributes are expressed.

\paragraph{General Domain.}
We collect paragraphs from mainstream news agencies, including The Economist, Fox News, CNN, and BBC, in the \texttt{Economics} and \texttt{Politics} domains to evaluate the methods' performance in general-purpose scenarios. In the Economics domain, the selected texts contain analyses with rich numerical data and fine-grained economic indicators, making it challenging for LLMs to identify and associate context-specific attribute–value pairs with the correct entities. For the Politics domain, all documents contain diverse entities whose attributes are scattered across sentences, posing challenges for extraction completeness. Successful extraction in this setting requires models to rely solely on contextual understanding to recognize entities and infer their corresponding attributes and values.

\subsection{Evaluation}

To comprehensively evaluate each method's ability to extract fine-grained information, we follow prior structured entity extraction and open information extraction work~\citep{dong-etal-2021-docoie, wu-etal-2024-structured_entity}, reporting \texttt{Precision}, \texttt{Recall}, and \texttt{F1} scores at the $\langle \text{entity}, \text{attribute}, \text{value} \rangle$ triplet level. To ensure high-quality ground truth annotations, we adopt a pooling-based evaluation strategy\cite{10.1145/3488560.3498377}: \textit{aggregate all extracted triplets across methods and have experienced annotators from each domain validate them to construct the reference set}. Full details on the evaluation criteria and annotation process are provided in Appendix~\ref{app:evaluation}.

\paragraph{Baselines.}
Since OpenESD requires contextual understanding to induce attributes from text—unlike traditional extractive information extraction tasks~\citep{nasar2021named, zhou2024relation_extraction}—we evaluate LLM-based approaches under both training-based and training-free settings. 

For the training-based setting, we report results from \textbf{Text2Triple}~\citep{jiang2025ras}, a 3B language model fine-tuned on a general-domain open triplet extraction dataset comprising 2 million instances curated using \texttt{Claude-Sonnet-3.5}. 

For training-free methods, we consider three prompting strategies: \textbf{Chain-of-Thought (CoT)} prompting~\citep{wei2022chain}, \textbf{Few-Shot} prompting~\citep{brown2020language}, and our proposed method \our. All three are evaluated using the following backbone models: \texttt{GPT-4o}~\citep{openai2024gpt4technicalreport}, \texttt{GPT-4o-mini}, and \texttt{Granite-8B}~\citep{granite2024granite}. Prompting templates are provided in Table~\ref{tab:prompting}.

\subsection{Main Results}
Table~\ref{tab: results} summarizes the performance of all evaluated methods across three domains: \texttt{Battery Science}, \texttt{Economics}, and \texttt{Politics} with three backbone models. We have the following observations: \our consistently achieves the highest F1 scores across all domains and backbone models, outperforming both \textbf{CoT} and \textbf{Few-Shot} prompting. This highlights the effectiveness and generalizability of \our in extracting accurate and comprehensive entity structures without relying on annotated data. However, we also observe that \our sometimes exhibits lower precision compared to other baselines. This may be because \our's enrichment module (cf. Section~\ref{triplet_enrichment}) not only recovers potentially missed extractions but also introduces noise into the results. We further analyze the contribution of each module of \our in Section~\ref{abaltion_analysis}.

Few-shot prompting generally improves performance, surpassing CoT in most cases in terms of precision, recall, and F1 score. This confirms the importance of in-context demonstrations in helping LLMs identify relevant attributes and values in open-schema settings. However, in the \texttt{Battery Science} domain, the improvement of few-shot prompting on recall is less pronounced, suggesting that in long-tail or highly specialized domains, few-shot examples may be insufficient for uncovering latent, context-dependent attributes—particularly when those attributes are nested within complex experimental conditions. These results highlight the benefit of \our's approach: abstracting attributes into coarse-grained representations to help LLMs uncover missing extractions, followed by a granularity refinement step to recover fine-grained contextual conditions.

While supervised fine-tuning can significantly enhance model performance on in-distribution data, such improvements often fail to generalize to unseen domains. In our experiments, Text2Triple~\citep{jiang2025ras}, a model fine-tuned on general domain, achieves strong performance in the \texttt{Politics} domain, with competitive scores in Precision, Recall, and F1. However, its effectiveness becomes less prominent in the \texttt{Economics} domain and drops substantially in the \texttt{Battery Science} domain. This degradation highlights the limited transferability of supervised approaches when faced with domain-specific or out-of-distribution contexts. In contrast, training-free methods, especially \our, demonstrate consistently robust performance across all domains, underscoring their adaptability and reliability in zero-shot settings.


\subsection{Ablation Analysis}\label{abaltion_analysis}
\begin{figure*}[t]
  \centering
  \includegraphics[width=\textwidth]{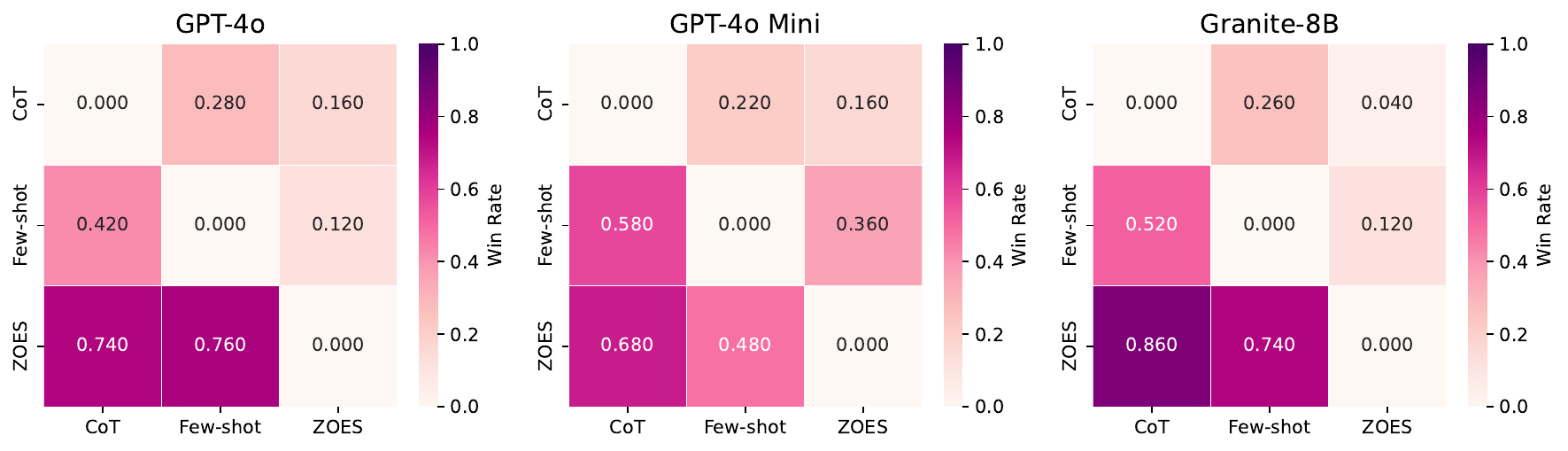}
    \caption{
        \textbf{Prompting-Based Extraction Coverage Win Rate} of different backbone models (\texttt{GPT-4o}, \texttt{GPT-4o Mini}, \texttt{Granite-8B}) using various prompting methods (\texttt{CoT}, \texttt{Few-Shot}, \texttt{ZOES}) in the Economics domain. 
        Each heat map shows the pairwise win rate between methods, where the value in row $i$, column $j$ represents the proportion of test instances for which method $i$ extracts more correct triplets than method $j$. 
        For example, with \texttt{GPT-4o}, \texttt{ZOES} outperforms \texttt{Chain-of-Thought} prompting in 74\% of instances (win rate = 0.740). 
    }
  \label{fig:win_rate_coverage}
\end{figure*}

To evaluate the contributions of \our's core components, we conduct ablation studies by removing two key modules: (1) Value-Anchored Enrichment (cf. Section~\ref{triplet_enrichment}) and (2) Mutual Dependency-Based Triplet Refinement (cf. Section~\ref{method: refinement}). We evaluate each variant using \texttt{GPT-4o} as the backbone model and report the results in Table~\ref{tab:ablation}.

\begin{table}[!ht]
\centering
\small 
\setlength{\tabcolsep}{4pt} 
\renewcommand{\arraystretch}{1.3} 
\begin{tabular}{l|ccc}
\toprule
\textbf{Method} & \textbf{Precision} & \textbf{Recall} & \textbf{F1} \\
\midrule
\our              & \textbf{0.8994} & \textbf{0.9104} & \textbf{0.9049} \\
w/o Enrich        & 0.8465 & 0.8758 & 0.8609 \\
w/o Refine        & 0.8143 & 0.8839 & 0.8477 \\
\bottomrule
\end{tabular}
\caption{Ablation results evaluated by \texttt{Precision}, \texttt{Recall}, and \texttt{F1} on the Finance domain using \texttt{GPT-4o} as the backbone.}
\label{tab:ablation}
\end{table}

As shown in Table~\ref{tab:ablation}, removing either component consistently degrades \our's performance, demonstrating the effectiveness of each module's design. Specifically, the Mutual Dependency-Based Triplet Refinement module is responsible for correcting potentially incorrect or incomplete extraction results. Removing this module noticeably reduces precision, as the model tends to include overgeneralized or ambiguous triplets that may have been introduced by the enrichment module.

These results also show that enrichment and refinement collaboratively enhance \our's performance: the enrichment module increases extraction coverage by discovering previously missed information, though it may also yield incomplete results due to the subtlety of certain implicitly mentioned attributes. Meanwhile, the refinement module helps detect and revise ambiguous or partial extractions, thereby improving the quality of enrichment.
\begin{table*}[t] 
\small
\centering
\setlength{\tabcolsep}{10pt} 
\renewcommand{\arraystretch}{1.5}

\begin{tabular}{>{\centering\arraybackslash}m{0.15\textwidth} | >{\centering\arraybackslash}m{0.75\textwidth}} 
\toprule
\textbf{Original Text} & Like other Japanese automakers, \textbf{Toyota} is very dependent on the US and \textbf{sold 2.3 million cars} in the country last year. About \textbf{one million} of those vehicles \textbf{were made in other countries}, many of them in Canada, Mexico and Japan. That could be a big problem for the company and automakers like \textbf{Subaru and Mazda}, with which \textbf{Toyota works closely}. But \textbf{Toyota, the world’s largest automaker}, is in a better position than other automakers. \textbf{It is profitable} and considered by analysts to be one of the best-run companies in the global auto industry... \\
\midrule
\textbf{Method} & \textbf{Extracted Entity Structures (Triplets)} \\
\midrule
\texttt{Few-shot} & 
\begin{tabular}[c]{c}
(Toyota, largest automaker, world), \\
(Toyota, annual car sales in US, 2.3 million), \\ 
(Toyota, profitability status, profitable), \\ 
(Toyota, reputation among analysts, one of the best-run companies in global auto industry) 
\end{tabular}\\
\midrule
\our & 
\begin{tabular}[c]{c}
(Toyota, Cars Sold in the US Last Year, 2.3M),\\
(Toyota, Close Collaborators, Subaru \& Mazda), \\ 
(Toyota, Cars manufactured Outside US, 1M),  \\
(Toyota, Position Among Automakers, World's Largest),\\  
(Toyota, Profitability, Profitable) 
\end{tabular}\\
\bottomrule
\end{tabular}
\caption{Comparison of entity extraction results between \our and the \texttt{few-shot} baseline using \texttt{Granite-8B}. The original text highlights crucial information regarding ``Toyota'' in \textbf{bold} as the reference for the extracted triplets.}
\label{tab:toyota-case}
\end{table*}
\subsection{Coverage Win Rate}
To assess extraction coverage across methods, we compute a coverage win rate for each backbone model (GPT-4o, GPT-4o Mini, and Granite-8B) under three prompting strategies (CoT, Few-Shot, and \our) on a per-document basis in the economics domain. The coverage win rate is calculated based on pairwise comparisons. The formal definition of the win rate is provided in Appendix \ref{win-rate}. As shown in Figure ~\ref{fig:win_rate_coverage}, \our consistently achieves higher win rates compared to both CoT and Few-Shot prompting across all models. Notably, these win-rate improvements indicate that the performance gains are not merely the result of a few information-dense documents, rather, \our demonstrates a robust capacity to extract comprehensive information consistently across diverse documents. The results from the figure~\ref{fig:win_rate_coverage} and table~\ref{tab: results} together demonstrate that even without training data, \our is capable of capturing more comprehensive information from diverse contexts, reinforcing its effectiveness in zero-shot open-schema entity structure discovery.
\subsection{Case Studies}
As shown in Table~\ref{tab:toyota-case}, \our produces more complete and contextually faithful extractions than Few-Shot prompting. First, \our captures more fine-grained and semantically rich attributes (e.g., ``Cars Sold in the US Last Year'', ``Close Collaborators'') compared to the relatively generic expressions extracted by Few-Shot (e.g., ``annual car sales in US''). This improvement stems from \our's mutual-dependency-based triplet refinement, which detects and refines ambiguous triplets. Second, \our demonstrates better coverage by identifying additional informative triplets that are absent in Few-Shot results (e.g., ``(Toyota, Cars manufactured
Outside US, 1M)'').  This is enabled by the value-anchored enrichment mechanism, which revisits the document to recover missing triplets under guided root attributes.
\paragraph{Observed Errors.}
While \our achieves superior recall and F1 scores, we observe a relatively lower precision in certain scenarios. Our analysis identifies two primary sources of these errors: 

\noindent\textbf{Overgeneralized Enrichment:} During the Value-Anchored Enrichment phase (Section \ref{triplet_enrichment}), \our utilizes coarse-grained root attributes to guide the discovery of additional value mentions. However, because these root attributes are intentionally broad, the model occasionally extracts general, sentence-level descriptions that are contextually relevant but lack the formal structure of a well-defined attribute. In our evaluation, such loosely structured outputs are penalized as incorrect, thereby impacting precision. 

\noindent\textbf{Ambiguous or Implicit Attributes:} \our relies on a value-anchored strategy, assuming each extracted value corresponds to a valid triplet. In cases where the underlying attribute is implicit and the context provides weak semantic cues, the model may struggle to infer a precise attribute name during the Triplets Granularity Refinement phase (Section \ref{method: refinement}). This can result in the generation of vague attributes (e.g., ``features,'' ``includes''), which are not ideal for structure discovery.

\section{Conclusions}
We introduce \our, a zero-shot, training-free framework for \textit{open-schema entity structure discovery} without relying on predefined schemas or annotated data. \our achieves high-quality entity structure extraction across both long-tail and general domains. Extensive experiments demonstrate that \our not only substantially improves the performance of smaller language models in a zero-shot setting, but also outperforms baselines across three diverse domains. Our findings suggest that explicitly structuring the entity discovery process rather than relying on static prompting alone offers a robust and principled approach to information extraction in long-tail, open-world scenarios. We believe \our is good experimental evidence for schema-free knowledge extraction with LLMs and provides a foundation for future research in context-grounded entity understanding.

\section*{Limitations}
This work introduces \our, a training-free zero-shot entity structure discovery method, and develops a dataset on three distinct domains to evaluate its performance against zero-shot and supervised baselines. We discuss the following limitations:

\paragraph{Computational Efficiency.} Although \our substantially improves LLM performance on open-schema entity structure extraction, it involves multiple rounds of generation, enrichment, and refinement. This pipeline process increases computational cost and inference time, which may hinder scalability. One potential research direction is to utilize \our extraction results as demonstrations for LLMs' few-shot learning on open-schema entity structure discovery.

\paragraph{Evaluation Metrics.} Our evaluation relies on human-annotated reference triplets and a weighted scoring function to assess the correctness and completeness of extracted structures. While this ensures high-quality assessment, the reliance on manual annotation can introduce subjectivity and may not scale efficiently to broader domains. Future work could explore more automated and domain-agnostic evaluation strategies to improve scalability and reproducibility.
\section*{Ethical Statement}
We uphold ethical principles throughout the design, development, and evaluation of \our. The dataset used in this work was curated with careful attention to exclude any personally identifiable or sensitive information. All documents included were collected in accordance with their respective licensing agreements and terms of use.

Human-annotated test data were collected with informed consent, following ethical research guidelines. To promote fairness and reduce potential bias, we curated a diverse dataset across three domains and verified that entity types and contextual structures were broadly representative.

\section*{Acknowledgments}
This work was supported in part by the IBM-Illinois Discovery Accelerator Institute (IIDAI), the AI Institute for Molecular Discovery, Synthetic Strategy, and Manufacturing: Molecule Maker Lab Institute (MMLI), funded by U.S. National Science Foundation under Awards No. 2019897 and 2505932, NSF IIS 25-37827, and the Institute for Geospatial Understanding through an Integrative Discovery Environment (I-GUIDE) by NSF under Award No. 2118329.  Any opinions, findings, and conclusions or recommendations expressed herein are those of the authors and do not necessarily represent the views, either expressed or implied, of DARPA or the U.S. Government.
 
\bibliography{ref}
\appendix
\clearpage
\appendix
\section{Evaluation}\label{app:evaluation}

\subsection{Evaluation Metrics}\label{app:evaluation-metrics}
Let each domain's dataset be \(\mathcal{D} = \{d_1, \ldots, d_{|\mathcal{D}|}\}\). For each document \(d \in \mathcal{D}\), let \(P_d\) denote the set of predicted triplets and \(G_d\) denote the set of ground-truth triplets.

Each predicted triplet \(t \in P_d\) is scored by human annotators using the following scoring function \(S(t)\), which measures the correctness and completeness of the extracted structure:
\begin{itemize}
    \item \(S(t) = 0\), if the triplet is \textbf{incorrect}, or if the entity is not of an \textit{interested type}.
    \item \(S(t) = 0.5\), if the triplet is \textbf{correct but incomplete}, e.g., the entity or value is only partially captured.
    \item \(S(t) = 1\), if the triplet is both \textbf{correct and complete}, with all components (entity, attribute, value) accurately captured.
\end{itemize}

To evaluate overall performance, we aggregate the scores across all documents. Define:
\[
P = \bigcup_{d=1}^{|\mathcal{D}|} P_d \quad \text{and} \quad G = \bigcup_{d=1}^{|\mathcal{D}|} G_d.
\]
We compute the evaluation metrics as:
\begin{subequations} \label{eq:eval_metrics}
\begin{align}
\mathrm{Precision} &= \dfrac{\sum\limits_{d=1}^{|\mathcal{D}|} \sum\limits_{t \in P_d} S(t)}{\sum\limits_{d=1}^{|\mathcal{D}|} |P_d|} \label{eq:precision} \\
\mathrm{Recall} &= \dfrac{\sum\limits_{d=1}^{|\mathcal{D}|} \sum\limits_{t \in P_d} S(t)}{\sum\limits_{d=1}^{|\mathcal{D}|} |G_d|} \label{eq:recall} \\
\mathrm{F1} &= 2 \cdot \dfrac{\mathrm{Precision} \cdot \mathrm{Recall}}{\mathrm{Precision} + \mathrm{Recall}} \label{eq:f1}
\end{align}
\end{subequations}

\subsection{Extraction Coverage Win-Rate}\label{win-rate}

To assess extraction coverage, we define the \textbf{Coverage Win Rate} based on pairwise comparisons. Let $M_i$ and $M_j$ be two prompting methods. For each document $d \in \mathcal{D}$, two annotators independently judge the results. Let $A_k(M_i, M_j, d) \in \{0, 1\}$ be the judgment of annotator $k$, where 1 signifies $M_i$ is more complete and informative than $M_j$.

A win is recorded only if both annotators reach a consensus. The win rate $W_{i,j}$ is defined as:

\begin{equation}
\begin{aligned}
W_{i,j} = \frac{1}{n} \sum_{d \in \mathcal{D}} \mathbb{I} \Big( & A_1(M_i, M_j, d) = 1 \\
& \land A_2(M_i, M_j, d) = 1 \Big)
\end{aligned}
\end{equation}

where $n$ is the total number of documents. If consensus is not reached, the comparison is marked as a tie.

\subsection{Human Annotation Protocol}
To ensure rigorous evaluation, we divided the annotation task into two teams based on domain expertise:
\paragraph{Battery Science Domain.} Two domain-expert researchers with Ph.D. degrees in science fields were recruited.:
\begin{itemize}
    \item One annotator collected all baseline outputs and corrected extraction errors to construct the ground-truth triplets.
    \item The another annotator independently received anonymized extraction results from each method and judged them as \textit{correct}, \textit{partially correct}, or \textit{incorrect} using the scoring rubric.
\end{itemize}
\paragraph{General Domain.} Three annotators participated:
\begin{itemize}
    \item A master's and an undergraduate student in computer science collaboratively constructed ground-truth triplets from model outputs, following the same procedure.
    \item A third annotator (a senior undergraduate student) independently evaluated the model predictions in a blind review setting using the scoring function.
\end{itemize}

\noindent This process ensures that the evaluation is both context-sensitive. 
We conducted an inter-annotator agreement analysis, which yielded an overall 
$\kappa$ score of $0.79$, indicating substantial agreement among the annotators.

\section{Prompting Templates \& Pseudo code of \our}
Table~\ref{tab:prompting} lists all prompting templates used in this study. For completeness, we also include the pseudocode of \our in Algorithm~\ref{alg:zoes}.
\begin{table*}[t]
\small
\centering
\begin{tabular}{@{}p{0.22\textwidth} p{0.80\textwidth}@{}}
\toprule
\textbf{Prompt Name} & \textbf{Prompt Template} \\
\midrule
0-shot Triplet Extraction &
You are an expert in information extraction. Extract all (entity, attribute, value) triplets from the document.  
Here is the Provided Document: [document] \\
\addlinespace

0-shot Root Attribute Induction &
You are a helpful information extraction assistant.  
Can you summarize a category name for the following values? \\
\addlinespace

0-shot Value Extraction &
You are a helpful information extraction assistant.  
Can you extract all values (exact text spans, with units) under [document] for each attribute in [root attribute]? \\
\addlinespace

0-shot Value-Guided Triplet Extraction &
You are an expert information extraction assistant.  
Given Document: [document] and value types, extract all values (exact text spans, with units) under each type. \\
\addlinespace

Mutual Dependency QA (Question Generation) &
You are a helpful question answering assistant.  
Given a <entity, attribute, value> triplet, generate three questions where each question asks for one component using only the other two as context.  
Do not infer or hallucinate new information.\\
\addlinespace

Mutual Dependency QA (Question Answering) &
You are a helpful question answering assistant.  
Please answer the following questions using answers extracted from the context.  
Context: [context]  
Question 1: Q\_entity  
Question 2: Q\_attribute  
Question 3: Q\_value \\
\addlinespace

Triplet Refinement &
There is a <entity, attribute, value> triplet extracted from the context.  
The original triplet may cause ambiguity due to an incomplete entity or a non-informative attribute.  
Refine the given triplet by extracting exact information from the context, such that the attribute is a clear property of the entity.  
Context: [context]  
Triplet: <entity, attribute, value> \\
\addlinespace

Entity Structure Construction &
For a given list of (entity, attribute, value) triplets and a context, merge triplets referring to the same entity into structured objects.  
Follow this format:  
{"entity name": {"attribute": "value", ...}, ...}  
Context: [document]  
Triplets: [triplets] \\
\addlinespace
Entity Type Filtration &
You are a helpful assistant. For a given entity with its attribute and values, can you decide whether the entity belongs to any given entity types based on the context. The given context is: [Context]. The given triplets are [Triplets]. The given entity types are: [Entity Type]. Response ``Yes'' or ``No''. \\
\addlinespace

Chain-of-Thought Triplet Extraction &
You are an expert in information extraction.  
Instructions:  
(1) Identify all precise entities of types in [T] that have associated characteristics.  
(2) For each entity, extract:  
- Entity: The name or title  
- Attribute: The key property  
- Value: The associated value (numerical, adjective, or noun phrase)  
Formatting:  
- Format exactly: [entity, attribute, value]  
Document: [document] \\
\addlinespace

Few-Shot Triplet Extraction &
You are an expert in information extraction.  
Instructions: same as Chain-of-Thought Triplet Extraction.  
In addition, you are given:  
Demonstrations: [Demonstrations]  
Document: [document] \\
\addlinespace
\bottomrule
\end{tabular}
\caption{Prompt templates used in this work. \texttt{[ ]} and \texttt{< >} denote placeholders.}
\label{tab:prompting}
\end{table*}

\vspace{-1em}
\begin{algorithm*}[t]
\caption{\our: Zero-Shot Open-Schema Entity Structure Discovery}
\label{alg:zoes}
\KwIn{Document $d$, Target entity types $\mathcal{T}$}
\KwOut{Structured entities $\mathcal{E}$}
\begin{multicols}{2}
\small
\textbf{Step 1: Triplet Candidates Extraction}\\
$T_{\text{init}} \leftarrow$ \texttt{LLM\_ZeroShotExtract}($d$) \\
$\mathcal{E}_{\text{emb}} \leftarrow \{f(t) \mid t \in T_{\text{init}}\}$ \tcp{Embed triplets}
$\mathcal{C} \leftarrow$ \texttt{AgglomerativeClustering}($\mathcal{E}_{\text{emb}}, \alpha$) \\
$\mathcal{R} \leftarrow \emptyset$ \\
\ForEach{$C_i \in \mathcal{C}$}{
  $r_i \leftarrow$ \texttt{LLM\_SummarizeAttributes}($C_i$) \\
  $\mathcal{R} \leftarrow \mathcal{R} \cup \{r_i\}$
}
$T_{\text{enrich}} \leftarrow T_{\text{init}}$ \\
\ForEach{$r \in \mathcal{R}$}{
  $\mathcal{V}_r \leftarrow$ \texttt{LLM\_ExtractValues}($r, d$) \\
  \ForEach{$v \in \mathcal{V}_r$}{
    $t_{\text{new}} \leftarrow$ \texttt{LLM\_InferTripletByValue}($v, r, d$) \\
    \If{$t_{\text{new}} \neq \emptyset$}{
      $T_{\text{enrich}} \leftarrow T_{\text{enrich}} \cup \{t_{\text{new}}\}$
    }
  }
}

\vspace{0.3em}
\textbf{Step 2: Triplet Granularity Refinement} \\
$T_{\text{refine}} \leftarrow \emptyset$ \\
\ForEach{$t = \langle e, a, v \rangle \in T_{\text{enrich}}$}{
  $is\_consistent \leftarrow$ \texttt{True} \\
  \ForEach{$c \in \{e, a, v\}$}{
    $q_c \leftarrow$ \texttt{GenerateQuestion}($t \setminus \{c\}$) \\
    $a_c \leftarrow$ \texttt{LLM\_Answer}($q_c, d$) \\
    \If{$a_c \neq c$}{
      $is\_consistent \leftarrow$ \texttt{False}; \textbf{break}
    }
  }
  \eIf{$is\_consistent$}{
    $T_{\text{refine}} \leftarrow T_{\text{refine}} \cup \{t\}$
  }{
    $t' \leftarrow$ \texttt{LLM\_RefineTriplet}($v, d$) \\
    \If{$t' \neq \emptyset$}{
      $T_{\text{refine}} \leftarrow T_{\text{refine}} \cup \{t'\}$
    }
  }
}

\vspace{0.3em}
\textbf{Step 3: Entity Structure Construction} \\
$\mathcal{E}_{\text{init}} \leftarrow$ \texttt{LLM\_ConstructEntities}($T_{\text{refine}}, d$) \\
$\mathcal{E} \leftarrow \emptyset$ \\
\ForEach{$e \in \mathcal{E}_{\text{init}}$}{
  \If{\texttt{LLM\_IsTypeMatch}($e, \mathcal{T}, d$)}{
    $\mathcal{E} \leftarrow \mathcal{E} \cup \{e\}$
  }
}
\Return{$\mathcal{E}$}
\end{multicols}
\end{algorithm*}

\end{document}